# Divergent Paths: Separating Homophilic and Heterophilic Learning for Enhanced Graph-level Representations

Lei Han
leih0003@e.ntu.edu.sg
Nanyang Technological University
Singapore

Xu Jiaxing
Nanyang Technological University
Singapore

Dong Xia
Nanyang Technological University
Singapore

Ke Yiping
Nanyang Technological University
Singapore

## Abstract

Graph Convolutional Networks (GCNs) are predominantly tailored for graphs displaying homophily, where similar nodes connect, but often fail on heterophilic graphs. The strategy of adopting distinct approaches to learn from homophilic and heterophilic components in node-level tasks has been widely discussed and proven effective both theoretically and experimentally. However, in graph-level tasks, research on this topic remains notably scarce. Addressing this gap, our research conducts an analysis on graphs with nodes' category ID available, distinguishing intra-category and inter-category components as embodiment of homophily and heterophily, respectively. We find while GCNs excel at extracting information within categories, they frequently capture noise from inter-category components. Consequently, it is crucial to employ distinct learning strategies for intra- and inter-category elements. To alleviate this problem, we separately learn the intra- and inter-category parts by a combination of an intra-category convolution (IntraNet) and an inter-category high-pass graph convolution (InterNet). Our IntraNet is supported by sophisticated graph preprocessing steps and a novel category-based graph readout function. For the InterNet, we utilize a high-pass filter to amplify the node disparities, enhancing the recognition of details in the high-frequency components. The proposed approach, DivGNN, combines the IntraNet and InterNet with a gated mechanism and substantially improves classification performance on graph-level tasks, surpassing traditional GNN baselines in effectiveness.

## 1 Introduction

Graph Convolutional Networks (GCNs) [16] are effective in graph-based data analysis, leveraging the homophily property [19] where similar nodes often connect [1, 37]. Due to their emphasis on aggregating and smoothing local feature, GCNs excel at learning from homophilic components across various networks such as social networks [16] and biological networks [9]. However, this characteristic limits their effectiveness in handling heterophilic components, where different types of nodes are connected [37].

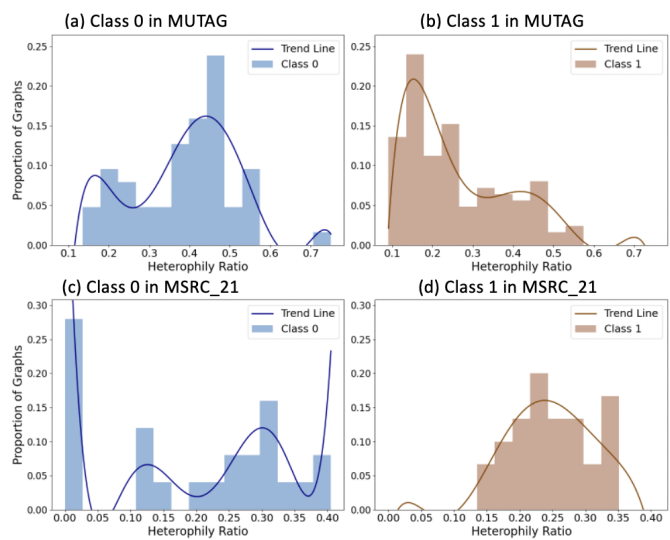

Figure 1: Heterophily ratio distribution of different classes in MUTAG [6] and MSRC_21 [29] dataset.

Table 1: Classification accuracy of GCNs on graphs with and without the heterophilic part

| Datasets | Whole Graph | w/o $\mathcal{E}^{hetero}$ |
| --- | --- | --- |
| DD [21] | 70.70±4.35 | 75.89±4.16 |
| MUTAG [6] | 82.40±5.95 | 81.93±8.22 |
| PTC_MM [12] | 63.36±6.29 | 66.05±9.09 |
| MSRC_21 [29] | 90.40±3.76 | 91.40±3.77 |

Therefore, to improve GCNs' adaptation on both homophily and heterophily, a series of node-level studies [5, 26, 30, 36], separating the learning for homophily and heterophily, have shown promising results both experimentally and theoretically. The spatial-based methods such as ego-neighbor separation [3, 7, 20, 32, 36] show effectiveness in addressing heterophily. They extract and differentiate the aggregated embeddings of the nodes themselves and their neighbors [37]. Usually, high and low frequency components relate to the graph's structural characteristics. Low frequency components, indicative of homophilic information, group similar or connected nodes to facilitate learning of local consistency and community traits [1, 17, 24, 30]. Conversely, high frequency components, which represent heterophilic information, highlight node differences and irregular connections, capturing anomalies and edge structures within the graph [30]. Naturally, the spectrum-based methods [2, 18, 30, 34] optimize the adaptation of GCNs on both low and high frequency components.

However, in graph-level tasks, research on homophily and heterophily remains very limited. The categorical labels for nodes are overlooked, such as atom types in chemical molecules [9] and segments in image graphs [29]. Naturally, subgraphs with intra-category connections align with homophily, while the subgraphs with inter-category connections correspond to heterophily. However, it is uncertain whether the node-level findings can be extended to the graph level. Specifically, are GCNs more effective for intra-category parts than inter-category parts? Are the separation strategies for intra- and inter-category components are effective for graph-level tasks?

To to answer these two questions, we first analyze the heterophily ratios' distribution of graphs across different ground truth classes, as illustrated in Figure 1. We define heterophilic edges connect nodes from different categories. The heterophily ratio is calculated by the proportion of heterophilic edges to all edges within a graph. Our findings indicate significant variations in the distribution of heterophilic ratios across different classes, highlighting the critical role of learning heterophilic information for graph-level classification tasks. It should be noted that, although MSRC_21 [29] is a multi-class dataset, we focused our analysis on classes 0 and 1 for plotting purposes.

We also conduct some comparative graph-level classification experiments with some generally used datasets [6, 29]. To compare with the original graph, we remove the heterophilic edges to create homophilic versions of the graphs. The results in Table 1 show that GCNs on homophilic graphs (w/o $\mathcal{E}^{hetero}$) even often outperform those on the original graphs (w/ $\mathcal{E}^{hetero}$), indicating that GCNs pick up noise from the heterophilic components during the learning process. This finding suggests the combined learning of homophily and heterophily may lead to sub-optimal results. Thus, we need more targeted approaches to address each component, thereby enhancing the adaptability of GCNs.

Consequently, we design a divergent-path learning approach called DivGNN, treating intra- and inter-category components separately. DivGNN aims to optimize the graph-level tasks learning by effectively addressing the distinct characteristics of homophily and heterophily. For learning the homophilic parts, we introduce a featureless intra-category convolution technique (IntraNet), which performs graph convolution operations within each category. Unlike traditional methods, IntraNet reduces the reliance on node features, thereby allowing for a greater focus on structural learning. IntraNet also includes a customized preprocessing procedure and a novel categorical readout function to better manage and utilize each category. For learning the heterophilic parts, we utilize a high-pass filter convolution [30], called InterNet, to emphasize the differences between nodes. Finally, our DivGNN, integrating IntraNet and InterNet through a gated sum approach, demonstrates superior efficacy over traditional GNN baselines.

Our main contributions can be summarized as follows:
- **Divergent paths for learning homophilic and heterophilic parts** We investigate the importance of distinct approaches to analyze and handle homophilic and heterophilic parts in graph-level tasks. We propose DivGNN, a model that integrates the learning for homophilic and heterophilic parts, which shows improved performance over traditional GNN baselines across a variety of datasets.
- **Intra-category convolution (IntraNet)** We propose a featureless intra-category convolution technique, without the dependence on node features, for homophilic parts graph-level representation learning. This approach includes a specific preprocessing technique and a novel category-based graph readout function.
- **Inter-category convolution with high-pass filter (InterNet)** We employ a high-pass filter for inter-category convolution, which emphasizes the heterophilic parts of graphs. Integrating the effective InterNet and IntraNet yields our comprehensive framework DivGNN.

## 2 Preliminaries
## 2.1 Spatial Domain

Consider an undirected graph $G = (\mathcal{V}, \mathcal{E})$, with $\mathcal{V}$ as the set of nodes and $\mathcal{E}$ as the set of edges. This study focuses on graphs characterized by categorical node labels, a common feature in various applications. For example, in molecular datasets [6], each atom's type is the node's category, while in image segmentation tasks [29], super-pixels are nodes with unique object IDs as categories. Normally, for these graphs, the initial nodes' feature is the one hot transformation of nodes' labels. In our graph model, each node $v \in \mathcal{V}$ is associated with a categorical label $c_v$ that belongs to a set $C = \{c^1, c^2, ..., c^k\}$, where $k$ denotes the total number of node categories.

**Homophily and Heterophily** In node-level task the homophily ratio $\alpha_v$ of node $v$, as defined in Pei et al. [26], quantifies the extent of label agreement among the neighbors of a node $v$ within a graph. This ratio is formally expressed as:

$$\alpha_v = \frac{\sum_{u \in N_v} 1(c_u = c_v)}{|N_v|}, \quad (1)$$

where $N_v$ denotes the set of neighbors of $v$, $c_u$ and $c_v$ represents the labels of nodes $u$ and $v$, respectively. A higher $\alpha_v$ value indicates a greater proportion of neighboring nodes share the same label as node $v$.

In graph-level tasks, we define the homophilic and heterophilic parts based on edges: we define the homophilic part subgraph $G^{homo} = (\mathcal{V}, \mathcal{E}^{homo})$, wherein $\mathcal{E}^{homo}$ denotes the set $\{e_{ij} \in \mathcal{E} \mid c_i = c_j\}$. While, the heterophilic part subgraph $G^{hetero} = (\mathcal{V}^{hetero}, \mathcal{E}^{hetero})$ only contains edges connected nodes with different labels, its edge set is $\mathcal{E}^{hetero} = \{e_{ij} \in \mathcal{E} \mid c_i \neq c_j\}$. The node set for the heterophilic part $\mathcal{V}^{hetero}$ consists of nodes in $\mathcal{V}$ excluding those with a homophily ratio of 1. Notably, while the edge sets $\mathcal{E}^{homo}$ and $\mathcal{E}^{hetero}$ are disjoint, $V^{hetero}$ is a subset of $\mathcal{V}$.

Moreover, in graph-level tasks, the heterophily ratio $\gamma$ of a graph dataset $\mathcal{G}$ is calculated as the average proportion of edges within the heterophilic part ($G^{hetero}$) relative to the total number of edges $|\mathcal{E}|$ in each graph:

$$\gamma = \frac{1}{|\mathcal{G}|} \sum_{G_i \in \mathcal{G}} \frac{|\mathcal{E}_i^{hetero}|}{|\mathcal{E}_i|} \quad (2)$$

where $|\mathcal{G}|$ represents the total number of graphs in the dataset, $|\mathcal{E}_i^{hetero}|$ is the number of edges in the heterophilic part of $G_i$, and $|\mathcal{E}_i|$ is the total number of nodes in graph $G_i$.

**Graph-Level Representation Learning.** Graph-level representation learning [4] involves Message Passing Neural Networks (MPNNs) [31] and readout functions. MPNNs aggregate node information based on connectivity, capturing the local structure to updated nodes' features $\{\mathbf{h}_v | v \in \mathcal{V}\}$. A readout function then consolidates these features into a single vector $\mathbf{H}_G$ representing the entire graph:

$$\mathbf{H}_G^{(l)} = \text{Readout}(\mathbf{h}_v^{(l)}, v \in \mathcal{V}), \quad (3)$$

where Readout(·) is a function that aggregates node features into a single graph-level representation of the $l$-th layer. The subgraph representations of $G^{homo}$ and $G^{hetero}$ are computed as:

$$\mathbf{H}^{homo\,(l)} = \text{Readout}(\mathbf{h}_v^{(l)}, v \in \mathcal{V}) \quad (4)$$

$$\mathbf{H}^{hetero\,(l)} = \text{Readout}(\mathbf{h}_v^{(l)}, v \in \mathcal{V}^{hetero}) \quad (5)$$

Commonly used readout functions include these types: sum pooling, $\mathbf{h}_G^{(l)} = \sum_{v \in \mathcal{V}} \mathbf{h}_v^{(l)}$, aggregates the total influence of nodes, while mean pooling, $\mathbf{h}_G^{(l)} = \frac{1}{|\mathcal{V}|} \sum_{v \in \mathcal{V}} \mathbf{h}_v^{(l)}$, calculates the average impact to normalize the features. Max pooling, $\mathbf{h}_G^{(l)} = \max_{v \in \mathcal{V}} \mathbf{h}_v^{(l)}$, highlights the dominant features. These methods are pivotal for tasks like toxicity assessment and disease diagnosis, where capturing different aspects of the graph's structure is crucial for accurate classification.

Virtual node readout aggregates the graph information by a virtual node connected to all real nodes. During processing, features from all nodes are aggregated to the virtual node, transformed, and then redistributed, enhancing node feature updates and capturing global graph properties:

$$\mathbf{h}_{\text{virtual}} = \sigma\left(\sum_{v \in \mathcal{V}} \mathbf{W}_{\text{agg}} \mathbf{h}_v + \mathbf{b}\right),$$

$$\mathbf{h}_v = \mathbf{h}_v + \mathbf{W}_{\text{redist}} \mathbf{h}_{\text{virtual}},$$

where $\mathcal{V}$ represents the set of real nodes and $h_{\text{virtual}}$ is embedding of the virtual node. $\sigma$ is a non-linear activation function, and $\mathbf{W}_{\text{agg}}$, $\mathbf{W}_{\text{redist}}$ are trainable parameters for aggregation and redistribution, respectively.

## 2.2 Spectral Domain

Graph Convolutional Networks (GCNs) [16] leverage spectral properties of graphs to perform convolutions that fuse node features with graph topology. Central to GCNs is the Laplacian matrix $\mathbf{L} = \mathbf{D} - \mathbf{A}$, with $\mathbf{A}$ as the adjacency matrix and $\mathbf{D}$ as the diagonal degree matrix. The eigenvalues and eigenvectors of $\mathbf{L}$ describe the graph's frequencies, with lower eigenvalues indicating smoother, more uniform structural features.

*Connection with Spatial Approaches.* To perform graph convolution, most recently, Balciar et al. [1] connect spectral-based approaches and spatial-based approaches by a uniform formula, which is defined as

$$\mathbf{X}^{(l+1)} = \sigma\left(\sum_{k=1}^{K} C^{(k)} \mathbf{X}^{(l)} \mathbf{W}_{l,k}\right), \quad (6)$$

with the graph convolutional kernel set to:

$$C^{(k)} = \mathbf{U}\,\text{diag}(F_k(\lambda))\mathbf{U}^T, \quad (7)$$

where $F_k(\cdot)$ is the filter function, $\mathbf{U}$ and $\lambda$ denote the eigenvectors and the eigenvalues of the normalized graph Laplacian matrix $\hat{\mathbf{L}} = \mathbf{I} - \mathbf{D}^{-\frac{1}{2}} \mathbf{A} \mathbf{D}^{-\frac{1}{2}}$ respectively, $\mathbf{D}$ is a diagonal matrix with $\mathbf{D}_{ii} = \sum_j \mathbf{A}_{ij}$, $\mathbf{W}_{l,k}$ is a learnable parameter.

## 3 Methodology

Figure 2 illustrates the schematic of our method DivGNN. We will describe our algorithm in the following sequence. Initially, we address the preprocessing for our intra-category graph convolution (IntraNet), as depicted in Figure 2(a). This stage is to isolate the homophilic components and separate nodes into category-based group through strategic node replication with label adjustment and node reordering. Following preprocessing, we delve into the intra-category convolution for learning the representations of the category subgraphs (Figure 2(b)). Additionally, a dedicated category-based readout (Figure 2(c)) is employed to integrate and combine the subgraph representations into the uniform representations. Finally, we elucidate the incorporation of the high-pass filter (Figure 2(d)) convolution called InterNet to improve the handling of the heterophilic components. The embeddings from homophilic and heterophilic components are gated combined into the final graph-level representations.

### 3.1 Preprocessing

**Node Replication with Label Adjustment.** In application graph datasets, there are some subgraph structures with decisive impact on predictive outcomes in industrial applications, such as the ring structures in predicting chemical compound properties [11]. However, some structures are mixed with nodes from different labels, as shown in Figure 3(a) and (b). This makes the structure of these subgraphs cannot be learned completely from the intra-category convolution. Therefore, in the preprocessing step, we first employ a node replication and label adjustment mechanism to preserve essential substructures that might otherwise be disrupted. Specifically, a node $v$ should be replicated with an adjusted label, if it fulfills the following criteria:

- Neighbor Number: $v$ must have at least two neighbors. This prevents misclassification of nodes with insufficient local connections.
- Neighborhood Purity: All neighbors of $v$ must have the same category label, ensuring complete homogeneity in its immediate vicinity. This condition is essential as it confirms the dominance of a single class within the local structure surrounding $v$.
- Label Difference: There must be a disparity between the label of $v$ and the unanimous label of its neighbors. This can help

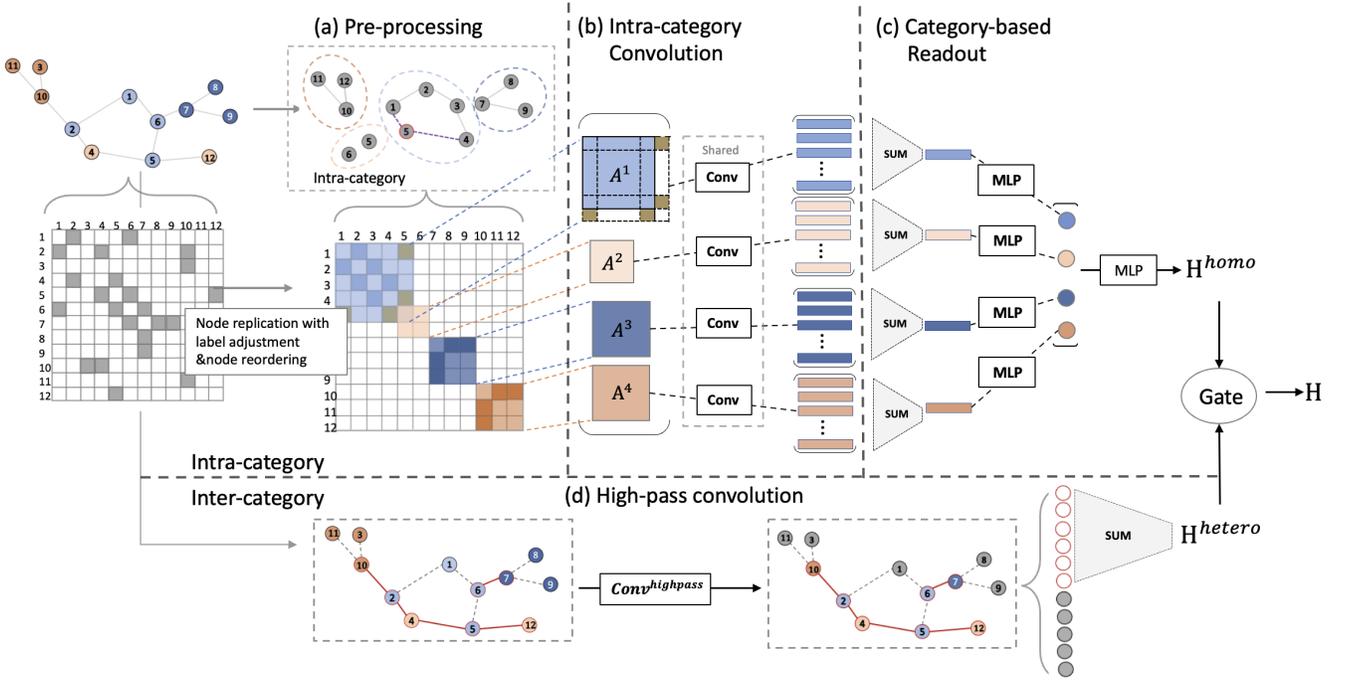

Figure 2: Schematic diagram of the DivGNN: (a) Initial graph preprocessing, including node replication with label adjustment and node reordering. The adjacency matrix is restructured into several intra-category adjacency matrices. (b) Within each category's subgraph, uniform node features enable featureless graph convolution, yielding specific subgraph embeddings for each category. (c) A category-based readout is then applied to derive the final homophilic part embedding. (d) High-pass convolution process focuses on encoding the heterophilic part. Ultimately, the embeddings from homophilic branch and heterophilic branch are gated and used as the prediction head.

identify the replication of $v$ is necessary to match the label of its surrounding nodes.

Upon satisfying these conditions, node $v$ is replicated and the its replication will be assigned a new label $\tilde{c}_v$ that inherits the common label of its neighbors $c_{n(v)}$ as shown in Figure 3(c)(d). This approach ensures that the presence of category impurities does not interfere with the learning of key substructures, such as ring structures. By effectively segregating and compensating for these impurities, our method maintains the integrity and focus for learning the essential structural features. The overall algorithm for node replication with label adjustment is outlined in the pseudo-code presented in Algorithm 1 in Appendix A.

**Node Reordering.** In order to prime the graph structure for more effective intra-category convolution, we reorder the nodes to cluster those with identical labels adjacent to each other within the adjacency matrix as shown in Figure 4. We define the horizontal index corresponding to each node's information in the graph's adjacency matrix as the node's order number. Normally, the initial nodes' orders in current graph datasets are typically arranged in a seemingly arbitrary order [21], as shown in Figure 4(a).

For our IntraNet, we reorder nodes such that nodes within the same category are assigned contiguous order numbers. A graph and its new adjacency matrix is reordered as depicted in Figure 4, while maintaining the structure of the homophilic part $G^{homo}$ unchanged.

The new adjacency matrix $\mathbf{A}^{homo}$ can be divided into several diagonal blocks $\{\mathbf{A}_1, \mathbf{A}_2, ..., \mathbf{A}_k\}$ shown in the colored matrix in Figure 4. $\mathbf{A}_1$ represents the adjacency matrix of subgraph of category 1, and $k$ is the number of categories. It is worth noting that if a node $v$ has been replicated, its replication should be arranged in the group of its neighbours' label $c_{n(v)}$. And $v$ should be a non-connection node in the group of $c_v$. This ensures that the information of the adjacency matrix $\mathbf{A}_{c_v}$ is complete. In this step, we obtain clean and well-organized homophilic information, facilitating our IntraNet more effectively on learning homophilic graphs.

### 3.2 Intra-category Graph Convolution

Given that GCNs are particularly effective at capturing low-frequency components [37], we develop IntraNet, a novel featureless intra-category graph convolution technique tailored for analyzing the structural information of these low-frequency elements. Low-frequency components typically involve nodes within the same category [30], which exhibit less variation and thus less signal frequency. Our targeted convolution operation, as demonstrated in Figure 2(b), is applied within nodes of the same category. We perform the convolution for each subgraph $G_i$:

$$\hat{\mathbf{A}}_i = \tilde{\mathbf{D}}_i^{-\frac{1}{2}} \tilde{\mathbf{A}}_i \tilde{\mathbf{D}}_i^{-\frac{1}{2}} \tag{8}$$

$$\mathbf{X}_i^{(l)} = \sigma\left(\hat{\mathbf{A}}_i \mathbf{X}_i^{(l-1)} \mathbf{W}^{(l-1)}\right) \tag{9}$$

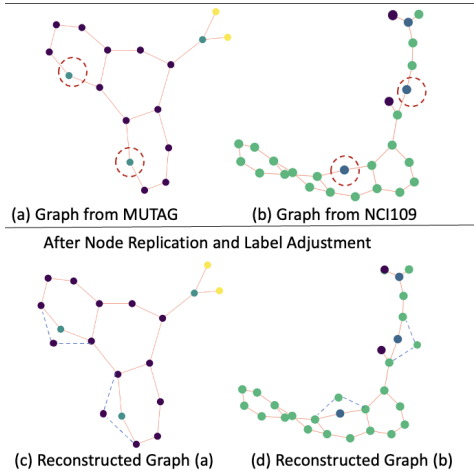

**Figure 3: Visualizations of original graphs and their modified versions post node replication and label adjustment: Nodes are colored based on their labels. Panels (a) and (b) depict the original graphs, highlighting mixed nodes within the core substructures. In panels (c) and (d), these selected nodes are replicated and the replicated versions are assigned new labels to match their neighboring nodes.**

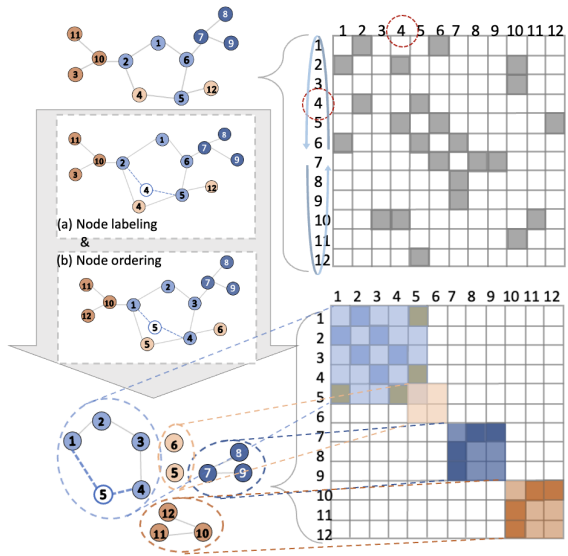

**Figure 4: Illustration of data preprocessing for intra-category graph convolution: node replication with label adjustment and node reordering.**

where $i \in \{1, 2, ..., k\}$ denotes the category ID, $\mathbf{A}_i$ and $\mathbf{X}_i$ denote the adjacency matrix and feature matrix of $G_i$, respectively.

Since the node features within subgraph $G_i$ are identical, the node feature matrix is not necessary within the IntraNet, solely relying on topology information. We replace the first layer's input node feature $\mathbf{X}_i$ of Equation (9) with a same-size identity matrix $\mathbf{I}_i$:

$$\mathbf{X}_i^0 = \mathbf{I}_i. \tag{10}$$

While for $l$ larger than 0, the intra-category convolution formula is the same as Equation (9). For each category channel, we do the intra-category convolution to update the subgraph representation set $\{\mathbf{X}_1^{(l)}, \mathbf{X}_2^{(l)}, ..., \mathbf{X}_k^{(l)}\}$ by Equation (10) and Equation (9).

IntraNet exclusively learns structural information for each specific category without using node feature information. Unlike traditional GCNs where categories are often entangled in a complex mix, our IntraNet has each category channel independent and categories do not interfere with each other (Figure 2(b)). This ensures that the structural learning is more precise. Moreover, the experimental results will demonstrate the effectiveness of our method: Even though IntraNet relies solely on the homophilic components, it achieves comparable or even better performance on most of datasets compared to GCNs that utilize the full graph input.

### 3.3 Category-based Readout

After getting $\mathbf{X}_2^{(l)}, \mathbf{X}_2^{(l)}, ..., \mathbf{X}_k^{(l)}\}$ from $l$ layers intra-category convolution, we develop a specific readout function called category-based readout. This novel approach is elaborated upon in Figure 2(c), and its main structure consists of an intra-category sum readout function and an inter-category concatenation function.

First, we place an intra-category sum readout followed with a Multi-Layer Perceptron (MLP) layer to achieve a low-dimensional subgraph embedding for each category subgraph:

$$\mathbf{h}_i' = \text{MLP}_1\Big(\sum_{v \in \mathcal{V}_i} \mathbf{h}_v^{(l)}\Big) \tag{11}$$

where $i = \{1, 2, ..., k\}$. However, when combining the representation of each category channel, traditional aggregation techniques, such as mean, maximum, and summation, might amalgamate the information and potentially lead to a loss of details of different category channels. To circumvent this issue, we concatenate the single category embeddings $\{\mathbf{h}_1', \mathbf{h}_2', ..., \mathbf{h}_k'\}$ together with each category occupied several digits, instead of combining them into the same digits. The final homophilic graph representations are computed by:

$$\mathbf{H}^{homo} = \text{MLP}_2(\text{concat}(\mathbf{h}_1', \mathbf{h}_2', ..., \mathbf{h}_k')) \tag{12}$$

In our category-based readout function, each category channel will be precisely learned and occupy an independent position in the final graph-level representation. Therefore, higher-level information from each channel is preserved to ensure a more robust representation. Furthermore, the efficacy and superior performance of our category-based readout method, as compared to conventional techniques, are verified by experimental results presented in the experimental section of our study.

### 3.4 The Encoding of Inter-Category Convolution

To leverage GCNs' strength in extracting low-frequency intra-category information, we decide to employ a high-pass filter to assist in extracting inter-category information. The high-pass filter computes the difference between the nodes' self-information and the neighbors' information, effectively highlighting the features of a node that are distinct from its neighbors. For the selection of the filter, we followed the approach used in AutoGCN [30], introducing two adjustable parameters to control the magnitude $p$ and the cut-off frequency $e$ of the filter function. The spectrum of the high-pass filter is designed to increase progressively, ensuring the effective separation of distinct node features. The high-pass linear filter function is defined as:

$$F_{\text{high}}(\lambda) = p(e\lambda + 1 - 2a). \tag{13}$$

After applying the high-pass convolution, as illustrated in Figure 2(d), the heterophilic aspects of the graph are accentuated and enhanced. According to Equation (6), we update the heterophilic embeddings in the $l$-th layer by employing the following high-pass convolution:

$$\mathbf{X}^{(l)} = \mathbf{C}_{\text{high}} \mathbf{X}^{(l-1)} \mathbf{W}^{(l-1)}, \tag{14}$$

where $\mathbf{C}_{\text{high}}$ is computed by Equation (7) and Equation (13). Then, the updated embeddings $\mathbf{X}^{(l)}$ is readout into the heterophilic graph-level representation:

$$\mathbf{H}^{\text{hetero}} = \text{SUM}(\mathbf{X}^{(l)}), \tag{15}$$

where $\text{SUM}(\cdot)$ is the sum pooling graph readout function. Both homophilic and heterophilic components are standardized and combined through a gating mechanism to form the final graph-level representation $\mathbf{H}^{\text{hetero}}$:

$$\mathbf{H} = \text{Gated}(\text{Standardize}(\mathbf{H}^{\text{homo}}), \text{Standardize}(\mathbf{H}^{\text{hetero}})), \tag{16}$$

where $\mathbf{H}^{\text{homo}}$ and $\mathbf{H}^{\text{hetero}}$ denote embeddings learned from the IntraNet and the high-pass graph convolution, respectively.

## 4 Experiment
### 4.1 Experimental Setup

We evaluate the effectiveness and versatility of our method across a range of graph datasets originating from diverse domains. Our datasets encompass both graphs of small molecules [12, 21] and large bioinformatics graphs [21] containing over 200 nodes, and their sizes vary significantly—from datasets contain hundreds of graphs to large ones with tens of thousands of graphs. Specifically, we utilize nine diverse public datasets from the TUDataset [21], spanning Chemistry, Bioinformatics, and Computer Vision. We also include large, imbalanced molecular datasets AQSOL [8] and ogbg-molhiv [14] for graph regression and graph classification tasks, respectively. The detailed dataset statistics are available in Appendix B.

### 4.2 Main Results

The experimental results displayed in the table 2 compellingly demonstrate the efficacy of DivGNN in graph-level classification tasks across a variety of chemistry-related and computer vision (CV) datasets. Overall, DivGNN achieves state-of-the-art (SOTA) results in 9 out of 10 datasets. Notably, it shows particularly significant improvements in certain datasets, consistently exceeding the second-best scores by approximately 3% to 4%. Specifically, the accuracy results of DivGNN on MUTAG, PTC_FM and PTC_FR surpass those on the run-up methods for 2.69%, 3.73% and 3.98%, respectively, highlighting the model's robustness and effectiveness. For CV datasets MSRC_9 and MSRC_21, our DivGNN also perform well, proving the efficacy of divergent-path learning for intra- and inter-pattern on CV datasets. This may also show the importance of inter-pattern components which can be seen as the borderline between each category-pattern as show in Figure 5(a)-(c). Moreover, for graph regression tasks, DivGNN also achieves the best performance among all baseline methods on the AQSOL [8] dataset. It should be noted that for the two large molecule datasets, AQSOL and ogbg-molhiv, we removed the node features but retained the node labels to meet the category setting requirements of our method; detailed explanations are provided in the appendix.

HGPSLPool [35] secures the second place on several datasets (4 out of 9), only outperformed by our method. Its success is attributed to a hierarchical learning process that assigns different levels to nodes, effectively enhancing performance in smaller graphs like the PTC series. In contrast, on larger datasets, this hierarchical structure might disrupt the balance between homophilic and heterophilic elements, limiting its effectiveness.

While our method has achieved substantial performance improvements, the time complexity of our approach remains comparable to that of a Graph Convolutional Network (GCN) with an equivalent number of layers. This aspect is further analyzed in the Appendix C.

### 4.3 Ablation Study and Analysis

***Effectiveness of The Homophilic and Heterophilic Branches***. To evaluate the effectiveness of our homophilic and heterophilic branches, we conducted an ablation study by individually removing each branch and comparing the graph classification accuracy with our DivGNN. The results in Table 3 demonstrate a decline in performance when either branch is removed, highlighting their significant contributions to the model. This finding indicates that both branches are vital for acquiring comprehensive graph information and generating high-quality graph-level representations.

When comparing IntraNet and DivGNN with our high-pass filter on the DD dataset, we observe a slight improvement. The DD dataset, characterized by its high heterophily ratio 0.93 (Table 5 in Appendix B), presents a challenge for enhanced performance gains. This dataset is notably complex, with each graph containing an average of around 284 nodes and spanning 89 categories (Table 5 in Appendix B). In such a complex graph structure, graph-level representation learning may compress a vast amount of information, leading to a blurring effect where detailed insights gained by sophisticated methods merge into generalized high-level embeddings. This blurring effect can cause intricate models to perform similarly to simpler ones, thus resulting in only modest improvements with the application of our high-pass filter in the DD dataset.

***The Comparison of Different Heterophilic Encoding Methods***. We explore various strategies for managing the heterophilic components of graphs, as detailed in Table 4. These include the commonly used ego-neighbor [37] GCNs, vanilla GCNs, and the high-pass filter method. The ego-neighbor GCNs are the commonly used heterophily-addressing method in node-level tasks, encoding each ego-embedding (i.e., a node's embedding) separately from the aggregated embeddings of its neighbors. The high-pass filter method is utilized in out DivGNN. As shown in Figure 4, The combination of IntraNet and high-pass convolution module demonstrates superior results, outperforming IntraNet with other heterophilic encoding methods. Furthermore, the combination of IntraNet with ego-neighbor GCNs consistently surpasses the version with vanilla GCNs, indicating that the integration of diverse neighborhood connections by ego-neighbor GCNs better supports the learning of heterophilic components. Lastly, the combination of IntraNet and GCNs for processing heterophilic parts yields the weakest results, primarily because GCNs are designed to operate under the conditions of homophily and struggle with heterophily [19].

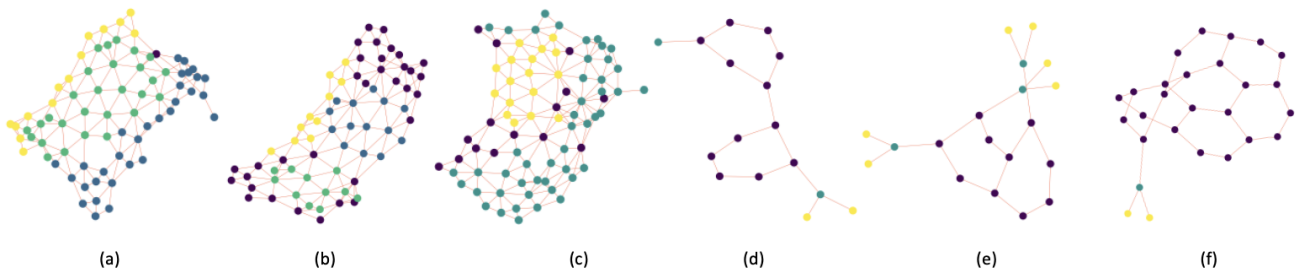

Figure 5: Visualization of six graphs from different domains:(a)-(c) are from the MSRC_21 Computer Vision dataset, while(d)-(f) are from the MUTAG chemical molecule dataset. Each graph distinguishes node labels using unique colors.

Table 2: Graph-level classification and regression results.

| Model | DD | MUTAG | PTC_FR [12] | PTC_MM | PTC_FM [12] | NCI1 [21] | NCI109 [21] | MSRC_9 [29] | MSRC_21 | AQSOL [8] | molhiv [14] |
|---|---|---|---|---|---|---|---|---|---|---|---|
| GIN [31] | 71.99±2.89 | 88.83±8.77 | 60.65±9.81 | 63.05±11.76 | 61.03±7.88 | **82.17±1.12** | **81.34±1.73** | 90.03±5.30 | 89.86±3.59 | 1.93 | 0.73 |
| GIN_vn | 74.18±3.63 | 86.08±9.80 | 61.24±6.36 | 59.53±7.35 | 55.30±8.38 | 76.92±2.19 | 79.84±1.87 | 90.49±4.29 | 90.60±3.70 | 1.98 | 0.71 |
| GCN [16] | 70.70±4.35 | 82.40±5.95 | 64.40±8.55 | 63.36±6.29 | 57.85±7.48 | 80.51±1.61 | 77.63±2.00 | 92.76±4.15 | 90.40±3.76 | **1.40** | 0.74 |
| GCN_vn | 73.48±6.35 | 75.99±7.79 | 65.82±10.93 | 59.80±5.71 | 60.54±4.41 | 69.22±1.97 | 75.84±2.04 | 89.58±3.57 | 88.10±3.53 | 2.26 | 0.70 |
| GAT [27] | 75.56±3.19 | 76.61±8.17 | 62.98±7.42 | 67.24±8.92 | 63.62±8.99 | 78.07±1.94 | 74.34±2.18 | **95.02±3.19** | 91.48±2.59 | 1.68 | 0.75 |
| NestedGIN [33] | 68.50±2.36 | 83.98±8.79 | 61.25±9.69 | 61.90±8.11 | 62.18±6.05 | 77.70±1.70 | 77.90±2.60 | 93.20±3.68 | 90.06±2.86 | 1.89 | 0.73 |
| MEWISPool [23] | 76.03±2.59 | 85.67±7.07 | 66.36±4.87 | 65.12±8.02 | 58.75±8.05 | 74.21±3.26 | 75.30±1.45 | 88.22±5.48 | 89.53±3.37 | 1.67 | 0.67 |
| HGPSLPool [35] | 71.25±3.25 | 79.82±8.49 | 67.52±5.62 | 67.50±6.29 | 63.62±8.99 | 79.26±1.44 | 75.83±1.98 | 93.18±4.19 | 92.02±2.50 | 1.56 | 0.74 |
| UGformer [22] | 76.65±3.44 | 76.05±8.81 | 66.37±6.89 | 64.83±8.07 | 60.15±8.24 | 68.82±1.24 | 68.36±2.45 | 89.11±5.09 | 88.83±3.48 | **1.40** | 0.72 |
| UGT [13] | 71.62±2.87 | 87.88±5.51 | 64.0±3.25 | 65.21±0.17 | 63.26±3.24 | 77.55±0.16 | 75.45±0.1.26 | 92.02±4.29 | 89.42±3.55 | 1.68 | 0.69 |
| CANON [10] | 77.00±2.90 | 85.61±8.25 | 67.51±2.75 | 65.13±4.91 | 61.91±5.66 | 65.42±2.69 | 64.09±3.49 | 94.09±2.91 | 91.13±3.52 | 1.55 | 0.70 |
| AutoGCN [30] | 73.17±2.91 | 86.70±6.78 | 61.25±9.08 | 64.60±5.91 | 60.74±6.55 | 80.05±1.17 | 79.90±1.96 | 91.84±4.47 | 89.90±3.91 | 1.54 | 0.71 |
| DivGNN | **77.94±3.10** | **91.52±6.34** | **71.50±6.91** | **68.16±8.34** | **67.35±10.35** | **82.17±1.91** | 79.98±1.87 | **95.02±3.19** | **92.83±3.76** | **1.40** | **0.76** |

[a] AQSOL [8] is utilized for a graph regression task, whereas the other datasets are employed for graph-level classification tasks.
[b] Suffix '_vn' indicates the model using a virtual node as a graph-level readout function.
[c] The best results are highlighted in bold font, and the runner-up results are underlined.
[d] The evaluation metric for tasks on AQSOL and ogbg-molhiv is MEA and AUC-ROC, respectively.

Table 3: Graph classification accuracy of IntraNet, InterNet and DivGNN.

| Model | DD | MUTAG | PTC_FR | PTC_MM | PTC_FM | NCI1 | NCI109 | MSRC_9 | MSRC_21 |
|---|---|---|---|---|---|---|---|---|---|
| DivGNN | **77.94±3.10** | **91.52±6.34** | **71.50±6.91** | **68.16±8.34** | **67.35±10.35** | **82.17±1.91** | 79.98±1.87 | **95.02±3.19** | **92.83±3.76** |
| IntraNet | 77.19±4.85 | 86.17±6.32 | 65.53±0.55 | 67.31±7.31 | 60.73±8.15 | 77.25±2.29 | 75.84±2.24 | 91.13±3.15 | 91.40±5.91 |
| InterNet | 73.17±2.91 | 87.72±8.62 | 66.97±11.45 | 65.14±5.89 | 60.74±6.55 | 79.90±1.96 | 80.71±1.90 | 93.18±5.47 | 89.84±4.47 |

***Effectiveness of Node Replication with Label Adjustment.***
We validate the impact of our node replication with label adjustment on our IntraNet through ablation study. Results depicted in Figure 6(a) illustrate the performance of the IntraNet with and without node relabeling in graph-level classification tasks. Generally, the inclusion of replication nodes consistently enhanced performance across most datasets. However, on CV datasets, MSRC_9 shows no change, while MSRC_21 experienced a slight decline, suggesting that this technique may be less critical for CV datasets. CV graphs consist of regular intra-category mesh patterns and intra-category patterns' boundaries (Figure 5(a)-(c)). For CV graphs, a node, qualified to be replicated, should be a single node located inside other category's pattern. Due to the relatively higher number of nodes in a CV intra-category pattern, the impact of a single replication node on learning this pattern is less significant.

Table 4: Graph classification accuracy of the combination of IntraNet and different learning methods for heterophilic parts.

| Model | DD | MUTAG | PTC_FR | PTC_MM | PTC_FM | NCI1 | NCI109 | MSRC_9 | MSRC_21 |
|---|---|---|---|---|---|---|---|---|---|
| DivGNN | **77.94±3.10** | **91.52±6.34** | **71.50±6.91** | **68.16±8.34** | **67.35±10.35** | **82.17±1.91** | **79.98±1.87** | **95.02±3.19** | **92.83±3.76** |
| _hetero(gcn) | 72.24±5.36 | 81.40±10.81 | 68.11±8.45 | 62.78±8.39 | 57.53±9.62 | 72.09±3.42 | 72.35±2.89 | 90.24±3.35 | 87.31±5.33 |
| _hetero(ego) | 75.39±3.80 | 83.57±9.77 | 68.12±3.96 | 64.87±3.56 | 54.69±6.44 | 72.70±3.92 | 76.04±1.48 | 88.46±3.96 | 91.40±3.77 |

[a] '_hetero(gcn)' indicates the gating combined version of IntraNet and vanilla GCNs heterophilic method.
[b] '_hetero(ego)' indicates the gating combined version of IntraNet and ego-neighbor separating GCNs heterophilic method.

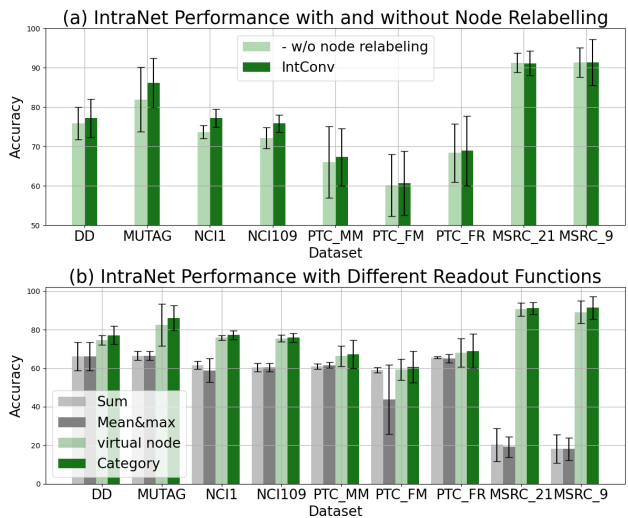

Figure 6: Comparison of different readout functions on IntraNet.

***Effectiveness of Category-based Readout Function.*** We conduct comparison experiments on our IntraNet model by using different readout functions, including traditional sum, mean/max, and virtual node readout functions. These experiments are performed in graph-level classification tasks, as shown in Figure 6(b). Our category-based readout function, highlighted in dark green, consistently outperforms these traditional methods across all datasets. This demonstrates its effectiveness and its capability to minimize information loss.

Compared to traditional sum and mean/max methods, our category-based readout function significantly enhances accuracy in CV datasets MSRC_9 and MSRC_21, boosting it from about 20% to over 91%.

This is because of the CV graphs' biased structural patterns within the same category (see Figure 5(a)-(c)), which cannot be distinguished after being combined by traditional methods. Specifically, the purple colored category constitutes two patterns in the graph in Figure 5(b), but it is only one node in the graph in Figure 5(a). In the final graph-level representations, our category-based readout helps preserve a position for each category channel and thus retains its uniqueness and differences. In contrast, chemical datasets do not have this dramatic margin, because specific categories often have specific structures (see Figure 5(d)-(f)), e.g. carbon atoms are usually in a ring-formed structure and chain-formed structure. Thus, the advantage of using a category-based readout over traditional methods is not as marked in chemical datasets as it is in CV datasets.

For the virtual node readout versions, they show robust performance across all datasets, slightly inferior to our category-based readout. The virtual node is integrated into each convolution layer of every category subgraph. It also can be seen as a form of category-based readout, which helps explain its effectiveness. However, the hierarchical plug-in of virtual nodes incurs a higher computational cost, especially for those graphs with more category numbers. We finally opted for the category-based readout as it can be easily implemented exclusively in the final layer of subgraph representation learning, while being simpler and without additional computation cost.

## 5 Conclusion

In conclusion, our study emphasizes the significance of separately addressing homophilic and heterophilic information at the graph level. For the homophilic part, with the support of our preprocessing techniques and category-based readout function, our IntraNet achieves comparable and even better results than those obtained by traditional GNN methods on complete graphs. Additionally, the introduction of a high-pass filter in our framework has been proven effective in enhancing the learning of heterophilic parts. This technology emphasizes the high-frequency components within graphs, which are crucial for extracting the inter-category information.

Table 5: The statistics of datasets used.

|  | DD | MUTAG | PTC_FM | PTC_FR | PTC_MM | NCI1 | NCI109 | MSRC_9 | MSRC_21 | AQSOL | molhiv |
|---|---|---|---|---|---|---|---|---|---|---|---|
| Avg.# of Nodes | 284.32 | 17.93 | 14.11 | 14.56 | 13.97 | 27.5 | 24.1 | 40.58 | 77.52 | 17.6 | 27.5 |
| # of Categories | 89 | 7 | 18 | 19 | 20 | 7 | 10 | 10 | 24 | 65 | 55 |
| # of Graphs | 1178 | 188 | 351 | 336 | 344 | 393 | 446 | 221 | 563 | 9823 | 41127 |
| Heterophily Ratio $\gamma$ | 0.93 | 0.27 | 0.39 | 0.39 | 0.39 | 0.37 | 0.37 | 0.31 | 0.25 | 0.34 | 0.34 |
| Domain | Bio | molecules | molecules | molecules | molecules | molecules | molecules | CV | CV | molecules | molecules |
| Eval | Acc | Acc | Acc | Acc | Acc | Acc | Acc | Acc | Acc | MAE ↓ | ROC_AUC ↑ |

[a] Bio denotes the domain of Bioinformatics. CV denotes the domain of Computer Vision.
[b] $\gamma$ denotes the average ratio of nodes existing in heterophilic parts.

## A  Psudo Code

To help understand the full algorithm of node replication with label adjustment, we present the pseudo-code as below:

---

**Algorithm 1:** Graph Reconstruction Algorithm via Node Replication with Label Adjustment

**Input** : Graph $G(\mathcal{V}, \mathcal{E})$
**Output**: Modified Graph $G'(\mathcal{V}', \mathcal{E}')$
**Init** : $G' \leftarrow G$

1 **for** $v \in \mathcal{V}$ **do**
2     **if** $|N(v)| \geq 2$
        **and** $\forall u, t \in N(v), c_u = c_t$
        **and** $c_v \neq c_{N(v)}$ **then**
3       $v' \leftarrow$ Replicate($v$);
4       SetLabel($v', c_{N(v)}$);
5       $\mathcal{V}' \leftarrow \mathcal{V} \cup \{v'\}$;
6       $\mathcal{E}' \leftarrow \mathcal{E} \cup \{(v', u) \mid u \in N(v)\}$;
7     **else**
8       **break**

   // Form the new graph with updated vertex and edge sets
9 $G' \leftarrow (\mathcal{V}', \mathcal{E}')$

---

## B  Experiments Settings

**Dataset selection.** We evaluate the effectiveness and generalizability of our method on graph datasets that include node categories. To initialize the node features, we utilize the one-hot transformation of category IDs. For a comprehensive assessment, we select 10 public datasets from TUDataset [21], encompassing three distinct domains: Chemistry, Bioinformatics, and Computer Vision. These datasets are for graph-level classification tasks. Additionally, we assess the performance of our model on large imbalanced molecular datasets, specifically AQSOL [8] and ogbg-molhiv [14] for graph regression and graph classification tasks, respectively. Detailed dataset statistics can be found in Table 5 in Appendix B.

**Baselines.** We benchmark our approach against several key baselines: foundational GCN [16] and GIN [31], both enhanced with virtual node readout for improved global information capture, and NestedGIN [33], which incorporates shortest path enhancements. For self-attention methods, we include GAT [27], which dynamically adjusts weights based on neighbor influences, UGformer [22], which integrates node features and connectivity in its weighting mechanism, and UGT [13], which encodes both local and long-range connectivity. Additionally, we compare our approach against advanced graph pooling methods like MEWISPool [23] and HGPSLPool [35]. Moreover, AutoGCN [30] integrates high-pass filter to highlight subtle graph features and enhance feature extraction.

**Node Feature Pre-processing for Large Datasets.** To evaluate the robustness of DivGNN across extensive datasets, we analyze its performance on the large-scale chemistry datasets, ogbg and AQSOL. These datasets feature a multitude of initial node attributes; however, to maintain fairness, only the atom type feature is utilized. Consistency is ensured by applying identical pre-processing steps to our method and all comparison baselines. For the Tudatasets [21], experiments are conducted using a 10-fold cross-validation scheme. For the ogbg and AQSOL datasets, which are equipped with predefined data splits, experiments are carried out using these established splits.

**Model and Training Settings.** In our experiments, we use an end-to-end training approach with the Adam optimizer [15], implemented in PyTorch [25] and utilizing the Deep Graph Library [28] for standardized graph data processing. Computations are performed on high-performance workstations with Intel(R) Core(TM) i9-10940X CPUs and NVIDIA GeForce GTX 3090 GPUs. Our models are trained with a batch size of 50 and an initial learning rate of 0.0007, halved periodically. We validate our approach on TUDataset [21] using 10-fold cross-validation, the same validation method applied to all baseline models.

## C  Time complexity

DivGNN has two parallel pipelines to learn the homophilic part and the heterophilic part, respectively. For the homophilic branch, the time complexity is $O(|\mathcal{E}_{homo}| \cdot F_{in} + N_v \cdot F_{in} \cdot F_{out})$, where $F_{in}$ and $F_{out}$ denotes input and output node feature dimensions. For the heterophily branch, the time complexity is with the same magnitude as GCN [16] $O(|\mathcal{E}| \cdot F_{in} + N_v \cdot F_{in} \cdot F_{out})$. Therefore, the overall time complexity of DivGNN is $O(|\mathcal{E}| \cdot F_{in} + N_v \cdot F_{in} \cdot F_{out})$.